\documentclass[letterpaper, 10 pt, conference]{ieeeconf}  
\IEEEoverridecommandlockouts                              
\usepackage{amsmath}
\usepackage{hyperref}
\usepackage{amsfonts}
\usepackage{amssymb}
\usepackage{dsfont}
\usepackage{graphicx}
\usepackage{wrapfig}
\usepackage{subcaption}
\usepackage[breakwords]{truncate}
\usepackage{nicefrac}       
\usepackage[ruled]{algorithm2e}
\usepackage{paralist}
\usepackage{color}
\usepackage[figuresleft]{rotating}
\usepackage{bbm}

\title{\LARGE \bf Self-Supervised Sim-to-Real Adaptation for\hspace{-0.2mm} Visual Robotic Manipulation}

\author{Rae Jeong, Yusuf Aytar, David Khosid, Yuxiang Zhou, Jackie Kay, Thomas Lampe, \\  Konstantinos Bousmalis, Francesco Nori
\thanks{Authors are with DeepMind London, UK. \{raejeong, yusufaytar, dkhosid, yuxiangzhou, kayj, thomaslampe, konstantinos, fnori\}@google.com. Qualitative results can be found in our supplementary video: \url{https://youtu.be/pmLASU_MW_o}}
}

\begin{document}

\maketitle
\thispagestyle{empty}
\pagestyle{empty}

\begin{abstract}

Collecting and automatically obtaining reward signals from real robotic visual data for the purposes of training reinforcement learning algorithms can be quite challenging and time-consuming. Methods for utilizing unlabeled  data can have a huge potential to further accelerate robotic learning. We consider here the problem of performing manipulation tasks from pixels. In such tasks, choosing an appropriate state representation is crucial for planning and control. This is even more relevant with real images where noise, occlusions and resolution affect the accuracy and reliability of state estimation. In this work, we learn a latent state representation implicitly with deep reinforcement learning in simulation, and then adapt it to the real domain using unlabeled real robot data. We propose to do so by optimizing sequence-based self-supervised objectives. These exploit the temporal nature of robot experience, and can be common in both the simulated and real domains, without assuming any alignment of underlying states in simulated and unlabeled real images. We propose \textit{Contrastive Forward Dynamics} loss, which combines dynamics model learning with time-contrastive techniques.  The learned state representation that results from our methods can be used to robustly solve a manipulation task in simulation and to successfully transfer the learned skill on a real system. We demonstrate the effectiveness of our approaches by training a vision-based  reinforcement learning agent for cube stacking. Agents trained with our method, using only 5 hours of unlabeled real robot data for adaptation, shows a clear improvement over domain randomization, and standard visual domain adaptation techniques for sim-to-real transfer.

\end{abstract}
\begin{figure}[th]
\centering
\begin{subfigure}{0.486\textwidth}
  \centering
  
  \includegraphics[width=0.98\linewidth]{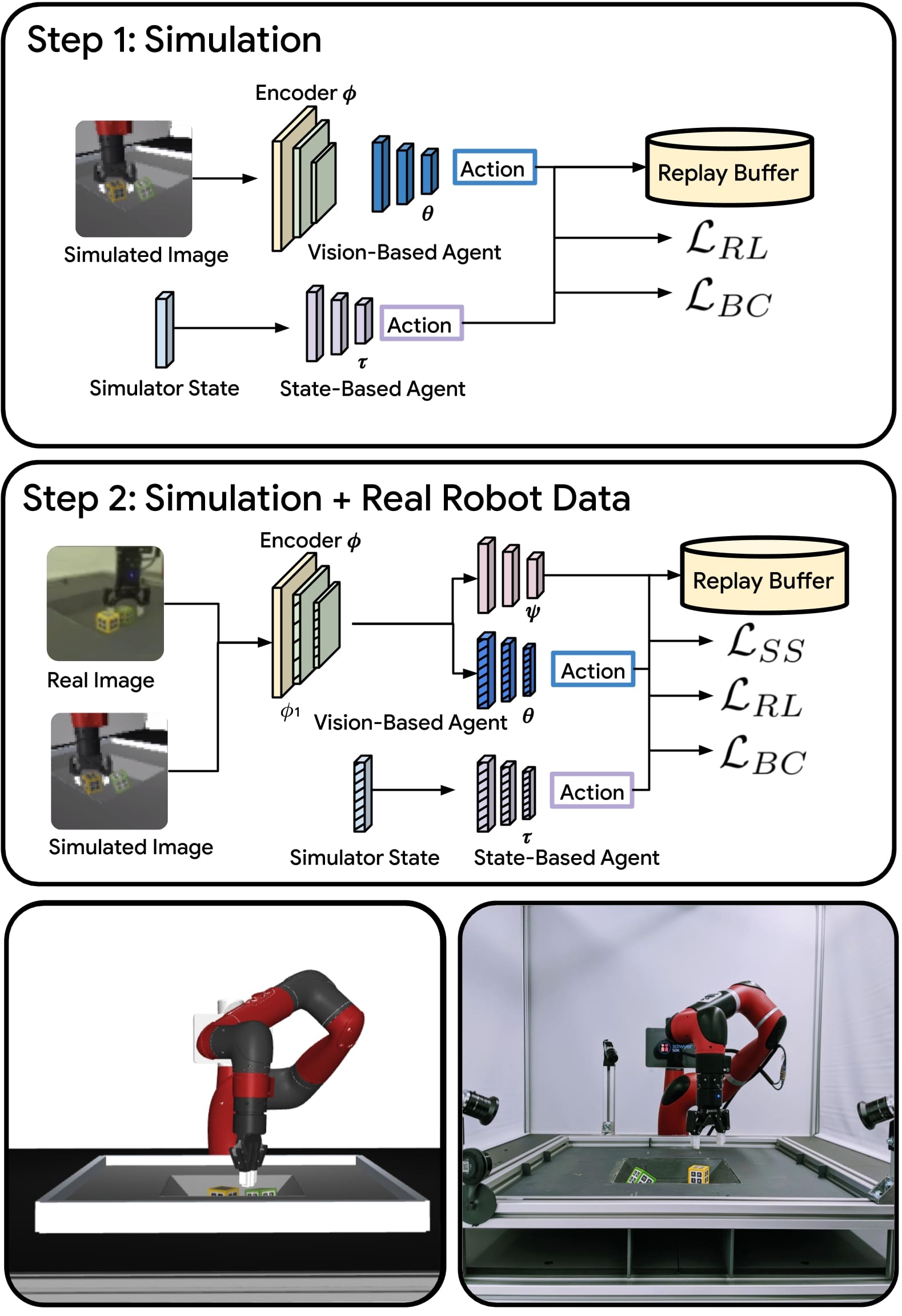}

  \label{fig:overview_pipeline}
\end{subfigure}

\caption{First step of our method trains a state-based and a vision-based agents in simulation. Using unlabeled real data we then perform domain adaptation with a sequence-based self-supervised objective, that are computable on both simulated and real data. The simulated and real robot setups we used are displayed at the bottom.}
\label{fig:overview}
\end{figure}

\section{Introduction}
\label{sec:introduction}
Learning-based approaches, and specifically the ones that utilize the recent advances of deep learning, have shown strong generalization capacity and the ability to learn relevant features for manipulation of real objects~\cite{bohg2014data,viereck2017learning,mahler2017dex,handeye_levine, qtopt}. These features can be used to avoid explicit object pose estimation~\cite{handbook_of_robotics} which is often inaccurate, even for known objects, in the presence of occlusions and noise. Furthermore, parameterization of the environment state with positions in $\mathbb{R}^3$ and rotations in $SO(3)$ is not necessarily the best state representation for every task.

Deep learning can provide task-relevant features and state representation directly from data. However, deep learning, and especially deep reinforcement learning (RL), requires a significant amount of data, which is a critical challenge for robotics~\cite{qtopt}. For this reason, sim-to-real transfer is an important area of research for vision-based robotic control as simulations offer an abundance of labeled data.

Pixel-based agents trained in simulation do not generalize naively to the real world. However, recent sim-to-real transfer techniques have shown significant promise in reducing real-world sample complexity. Such techniques either randomize the simulated environment in ways that help with generalization~\cite{dr_peng,rcan, openai_hand}, use domain adaptation~\cite{ stein2018genesis}, or both~\cite{da_konstantinos}. Our work falls in the scope of unsupervised domain adaptation techniques, i.e. methods that are able to utilize both labeled simulated and unlabeled real data. These have been successfully used both in computer vision~\cite{csurka2017domain} and in vision-based robot learning for manipulation~\cite{da_konstantinos} and locomotion~\cite{stein2018genesis}.

The contribution of our work is two-fold:
\textsl{(a)} we investigate the use of sequence-based self-supervision as a way to improve sim-to-real transfer; and \textsl{(b)} we develop \textit{contrastive forward dynamics} (CFD), a self-supervised objective to achieve that.
We propose a two-step procedure (see Fig. \ref{fig:overview}) for such sequence-based self-supervised adaptation. In the first step, we use the simulated environment to learn a policy that solves the task in simulation using synthetic images and proprioception as observations. In the second step, we use synthetic and unlabeled real image sequences to adapt the state representation to the real domain. Besides the task objective on the simulated images, this step also uses sequence-based self-supervision  as a way to provide a common objective for representation learning that applies in both simulation and reality without the need for paired or aligned data.
Our CFD objective additionally combines dynamics model learning with time-contrastive techniques to better utilize the structure of sequences in real robot data.

We demonstrate the effectiveness of our approach by training a vision-based cube stacking RL agent. Our agent interacts with the real world with 20Hz closed-loop Cartesian velocity control from vision which makes our method applicable to a large set of manipulation tasks. The cube stacking task also emphasizes the generality of our approach for long horizon  manipulation tasks. Most importantly, our method is able to make better use of the available unlabeled real world data resulting in higher stacking performance, compared to domain randomization~\cite{dr_tobin} and domain-adversarial neural networks~\cite{dann}.

\section{Related Work}
\label{sec:related}

\paragraph{Manipulation: challenges and approaches}It is well acknowledged that both planning and state estimation become challenging when performed in cluttered environments~\cite{billard2019trends}. During execution, continuously tracking the pose of manipulated objects becomes increasingly more difficult in presence of occlusions, often caused by the gripper itself. Surveys reveal that pose estimation is still an essential component in many approaches to grasping \cite[fig.~3-5-7]{bohg2014data}; proposed approaches rely on some sort of supervision, either in the form of model-based grasp quality measure~\cite{force_closure_1988,caging_servey, caging_servey2}, or in the form of heuristics for grasp stability~\cite[fig.~18-19]{bohg2014data}, or finally in the form of labelled data for learning~\cite[fig.~9]{bohg2014data}.

\paragraph{Sim-to-Real Transfer for Robotic Manipulation}

Sim-to-real transfer learning aims to bridge the gaps between simulation and reality, which consist of differences in the dynamics and observation models such as image rendering. Sim-to-real transfer techniques can be grouped by the amount and kind of real world data they use. Techniques like domain randomization~\cite{openai_hand, dr_tobin} focus on zero-shot transfer. Others are able to utilize real data in order to adapt to the real world via \textit{system identification} or \textit{domain adaptation.} Similar to system identification in classical control~\cite{sysid_emo}, recent techniques like SimOpt~\cite{simopt} utilize real data to learn policies that are robust under different transition dynamics.
Unsupervised domain adaptation~\cite{csurka2017domain} has been successfully used for sim-to-real transfer in vision-based robotic grasping~\cite{da_konstantinos}. Semi-supervised domain adaptation additionally utilizes any labeled data that might be available, as was done by~\cite{da_konstantinos}. In many ways, zero-shot transfer, system identification, domain adaptation--with or without labeled data in the real world--are complementary groups of techniques.

\paragraph{Cube Stacking Task}Recent work on efficient multi-task deep reinforcement learning~\cite{riedmiller2018learning} has shown the difficulty of cube stacking task even in simulated environments as the task requires several core abilities such as grasping, lifting and  precise placing. Sim-to-real method has also been applied for cube stacking task from vision where combination of domain randomization and imitation learning was used to perform zero-shot sim-to-real transfer of the cube stacking task~\cite{Zhu-RSS-18}. However, the resulting policy only obtained a success rate of
35\% over 20 trials in a limited number of configurations reconfirming the difficulty of the cube stacking task.

\paragraph{Unsupervised Domain Adaptation}Unsupervised domain adaptation techniques are either feature-based or pixel-based. Pixel-based adaptation is possible by changing the observations to match those from the real environment with image-based GANs~\cite{bousmalis2017unsupervised}. Feature-based adaptation is done either by learning a transformation over fixed simulated and real feature representations, as done by~\cite{caseiro2015beyond} or by learning a domain-invariant feature extractor, also represented by a neural network~\cite{ganin2016domain, bousmalis2016domain}. The latter has been shown to be more effective~\cite{bousmalis2016domain}, and we employ a feature-level domain adversarial method~\cite{ganin2016domain} as a baseline.

\paragraph{Sequence-based Self Supervision}

Sequence-based self-supervision is commonly applied for video representation learning, particularly making use of local~\cite{fernando2017self} and global~\cite{wei2018learning} temporal structures. Time-contrastive networks (TCN)~\cite{tcn} utilize two temporally synchronous camera views to learn view-independent high-level representations. By predicting temporal distance between frames, Aytar et al. \cite{aytar2018playing} learn a representation that can handle small domain gaps (i.e.\ color changes and video artifacts) for the purpose of imitating YouTube gameplays in an Atari environment. To the best of our knowledge, sequence-based self-supervision for handling large visual domain gaps in sim-to-real transfer for robotic learning have not been considered before.

\section{Our Method}
\label{sec:our_methods}

In this section, we provide the detailed description of our method for enabling sim-to-real transfer of visual robotic manipulation.
We propose a two stage training process. In the first stage, state-based and vision-based agents are trained simultaneously in simulation with domain randomization.

We then collect unlabeled robot data by executing the vision-based agent on the real robot. In the second stage, we perform self-supervised domain adaptation by tuning the visual perception module with the help of sequence-based self-supervised objectives optimized over simulation and real world data jointly.

Our method optimizes three main loss functions: \textsl{(a)} $\mathcal{L}_{RL}$
is the reinforcement learning (RL) objective optimized by the state-based and vision-based agents in simulation, \textsl{(b)} $\mathcal{L}_{BC}$ is the behavioral cloning loss utilized by the vision-based agent to speed up learning by imitating the state-based agent, and \textsl{(c)} $\mathcal{L}_{SS}$ is the sequence-based self-supervised objective optimized on both simulation and real robot data. The purpose of $\mathcal{L}_{SS}$ is to align the agent's perception of real and simulated visuals by solving a common objective using a shared encoder. 

Our system is composed of four main neural networks: \textsl{(a)} an image representation encoder with parameters $\boldsymbol{\phi} = \{\phi_i\}_i^L$ composed of $L$ layers which embeds any visual observation $o_t$ to a latent space as $z_t=\boldsymbol{\phi}(o_t)$, \textsl{(b)} a vision-based deep policy network with parameters $\boldsymbol{\theta}$ which combines the output of the visual encoder with the proprioceptive observations and outputs an action, \textsl{(c)} a state-based policy network with parameters $\boldsymbol{\tau}$ which takes the simulation state and outputs an action, and \textsl{(d)} a self-supervised objective network with parameters $\boldsymbol{\psi}$ which takes the encoded visual observation $z_t$ (and action $a_t$ if necessary) as input and directly computes the loss $\mathcal{L}_{SS}$. Fig. \ref{fig:overview} presents a visual description of these components. In the remainder of this section, we discuss the two stages of our method and present an objective for sequence-based self-supervision.

\subsection{First stage: Learning in simulation}
\label{sec:aai}
In this stage we train a state-based agent and a vision-based agent with a shared experience replay. Our goal is to speed up the learning process by leveraging the privileged information in simulation through the state-based agent, and distilling the learned skills into the vision-based agent using a shared replay buffer. Both of the agents are trained with an off-policy reinforcement learning objective, $\mathcal{L}_{RL}$. We use a state of the art continuous control RL algorithm, Maximum a Posteriori Policy Optimization (MPO)~\cite{mpo}, which uses an expectation-maximization-style policy optimization with an approximate off-policy policy evaluation algorithm. As shown in Fig. \ref{fig:overview}, the state-based agent has access to the simulator state, which allows it to learn much faster than the
vision-based agent that uses raw pixel observations. In essence, the state-based agent is an asymmetric behavior policy, which provides diverse and relevant data for reinforcement learning of the vision-based agent. This idea leverages the flexibility of off-policy RL, which has been shown to improve sample complexity in a single-domain setting~\cite{bic}. Additionally, we also utilize the behavioral cloning (BC) objective~\cite{bc} for the vision-based agent to imitate the state-based agent. $\mathcal{L}_{BC}$ provides reliable training and further improves sample efficiency in the learning process, as we show in Sect. \ref{sec:exp_results}. We additionally employ DDPGfD~\cite{ddpgfd} which injects human demonstrations to the replay buffer and \textit{asymmetric actor-critic} for our stacking experiments. Our final objective in the first stage can be written as follows:

\begin{equation}
\min_{\boldsymbol\phi,\boldsymbol\theta,\boldsymbol\tau} \quad  \mathcal{L}_{RL} + \mathcal{L}_{BC} \label{eq:aai} \\
\end{equation}

\begin{figure}[t]
\centering
\begin{subfigure}{0.486\textwidth}
  \centering
  
  \includegraphics[width=0.95\linewidth]{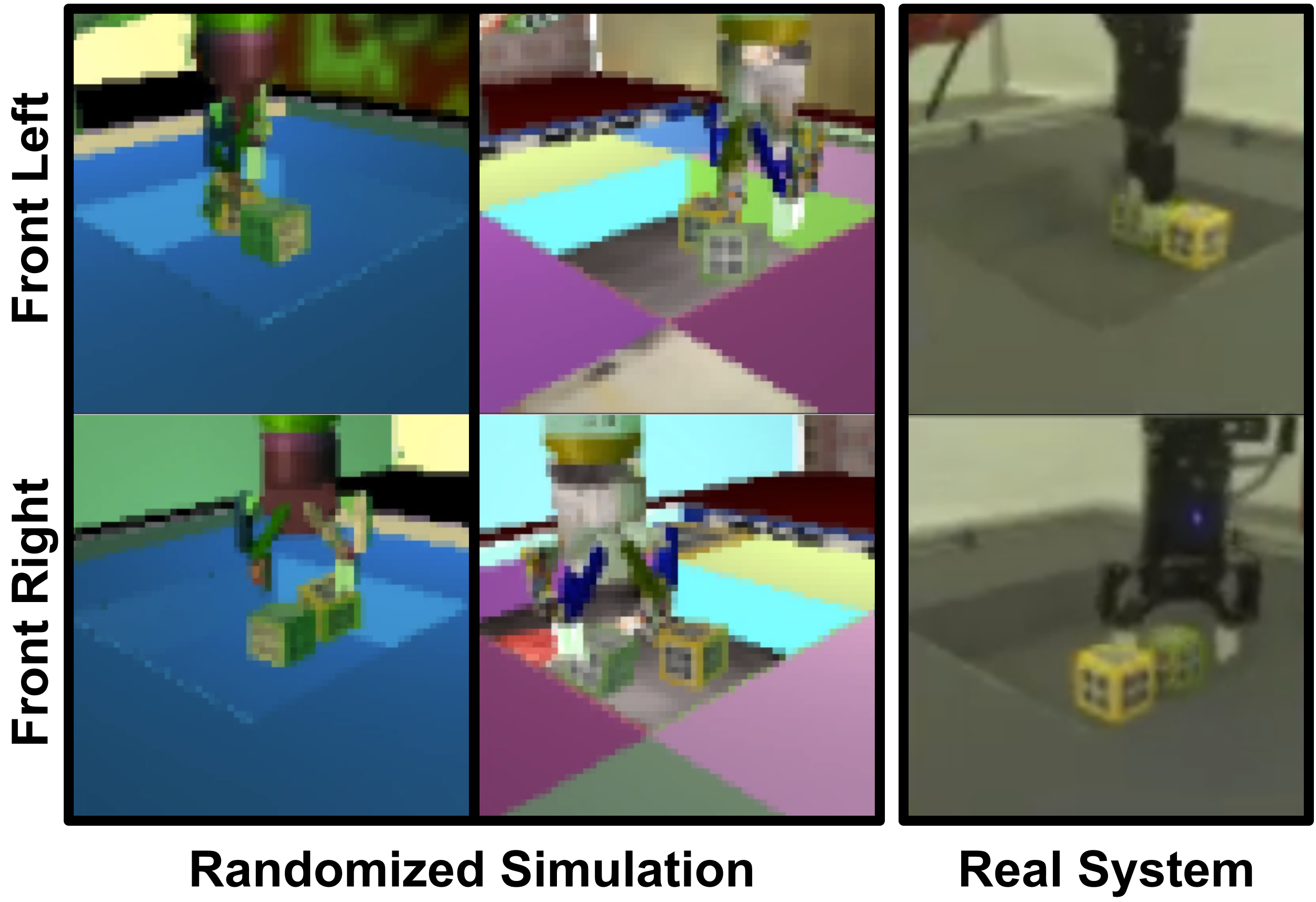}

  \label{fig:real_robot_front}
\end{subfigure}
\caption{Left and right pixel observations in both real and domain randomized simulated environments.}
\label{fig:stacking}
\end{figure}

\subsection{Second stage: Self-supervised sim-to-real adaptation}
\label{sec:selfsup}
Although our vision-based agent can perform reasonably well when transferred to the real robot, there is still significant room for improvement, mostly due to the large domain gap between simulation and the real robot. Our main objective in this stage is to mitigate the negative effects of the domain gap by utilizing the unlabeled robot data collected by our simulation-trained agent for domain adaptation. In addition to well-explored domain adversarial training~\cite{ganin2016domain}, which we present as a strong baseline, we investigate the use of sequence-based self-supervised objectives for sim-to-real domain adaptation.

\emph{Modality tuning}~\cite{aytar2017cross}, freezing the higher-level weights of a trained network and adapting only the initial layers for a new modality (or domain), is a method shown to successfully align multiple modalities (i.e. natural images, line drawings and text descriptions), though it requires class labels in all modalities. In our context, it would require rewards for the real-world data which we do not have. Instead, we utilize a self-supervised objective while performing modality tuning (i.e. simulation-to-reality adaptation) which can be readily applied both in simulation and reality. However, there is no guarantee that this alignment learned using a $\mathcal{L}_{SS}$ objective would indeed successfully transfer the vision-based policy from simulation to the real world. In fact, different $\mathcal{L}_{SS}$ objectives would result in different transfer performances. Finding a suitable $\mathcal{L}_{SS}$ objective for better transfer of the learned policy is of major importance as well.

In the context of our neural network architecture, while applying the modality tuning, we freeze the vision-based agent's policy network parameters $\boldsymbol\theta$ and the encoder parameters $\boldsymbol\phi$ except for the first layer $\phi_1$. This allows the system to adapt its visual perception to the real world without making major changes in the policy logic, which we expect to be encoded in the higher layers of the neural network. We also continue optimizing the $\mathcal{L}_{RL}$ and $\mathcal{L}_{BC}$ objectives along with $\mathcal{L}_{SS}$ to ensure that as $\phi_1$ is adapting itself to solve the $\mathcal{L}_{SS}$, it also maintains good performance for the manipulation task. In other words, $\phi_1$ is forced to adapt itself without compromising the performance of the vision-based agent. The final objective in the second stage is:

\begin{equation}
\min_{\phi_1,\boldsymbol\psi} \quad  \mathcal{L}_{RL} + \mathcal{L}_{BC}  + \mathcal{L}_{SS}\label{eq:ss} \\
\end{equation}

Due to its wide adoption in the robotics settings, we employ the Time-Contrastive Networks (TCN) \cite{tcn} objective for $\mathcal{L}_{SS}$ in our self-supervised sim-to-real adaptation method, though any other sequence-based self-supervised objective can also be used here. In the next subsection we introduce an alternative loss for $\mathcal{L}_{SS}$ which makes use of domain-specific properties of robotics, therefore potentially result in better transferable alignment.

\subsection{Contrastive Forward Dynamics}
\label{sec:cd}

Time-Contrastive Networks (TCN)~\cite{tcn}, which we use as a baseline, and other sequence-based self-supervision methods \cite{aytar2018playing,misra2016shuffle,oord2018representation}, mainly exploit the temporal structure of the observations. However, with robot data we also have physical dynamics of the real world probed by actions and perceived through observations. In this section we describe the \textit{contrastive forward dynamics} (CFD) objective, which is able to utilize both observations and actions by learning a forward dynamics model in a latent space. Essentially we are learning the latent transition dynamics of the environment which has strong connections to the model-based optimal control approaches~\cite{mpc}. Therefore we can expect that the alignment achieved through our CFD objective potentially better transfers the learned policy from simulation to real world. We formally define the CFD objective below.

Assume we are given a dataset of sequences where each sequence $s=\{(o_t,a_t)\}_t^T$ is of length $T$. $o_t$ denotes observations and $a_t$ denotes the actions at time $t$.
Any observation $o_t$ is embedded into a latent space as $z_t = \phi(o_t)$ through the encoder network $\phi$. Given a transition $(z_t, a_t, z_{t+1})$ in the latent space, the forward dynamics model predicts the next latent state as $\hat{z}_{t+1} = f(z_t,a_t)$ where $f$ is the prediction network. Instead of learning $f$ by minimizing the prediction error $||\hat{z}_{t+1}-z_{t+1}||$, which has a trivial solution achieved by setting the latents to zero, we minimize a contrastive prediction loss. A contrastive loss~\cite{chopra2005learning,hadsell2006dimensionality} takes pairs of examples as input and predicts whether the two elements in the pair are from the same class or not. It can also be implemented as a multi-class classification objective comparing one positive pair and multiple negative pairs \cite{sohn2016improved}, creating an embedding space by pushing representations from the same ``class'' together and ones from different ``classes'' apart. In our context, $(\hat{z}_i,z_i)$ is our positive pair and any other non-matching pairs $(\hat{z}_i,z_k)$ where $k \neq i$ are the negative pairs. With CFD, we solve such a multi-class classification problem by minimizing the cross-entropy loss for any given latent observation $z_i$ and its prediction $\hat{z}_i$ as follows:

\begin{equation}
\min_{\phi,f} \,\left\{ -\log \, \ \frac{e^{-||\hat{z}_i-z_i||}}{e^{-||\hat{z}_i-z_i||} + \sum_{k \neq i}e^{-||\hat{z}_i-z_k||}}\right\} \label{eq:cfd} \\
\end{equation}

\begin{figure}[t]
\centering
\begin{subfigure}{0.486\textwidth}
  \centering
  \includegraphics[width=0.9\linewidth]{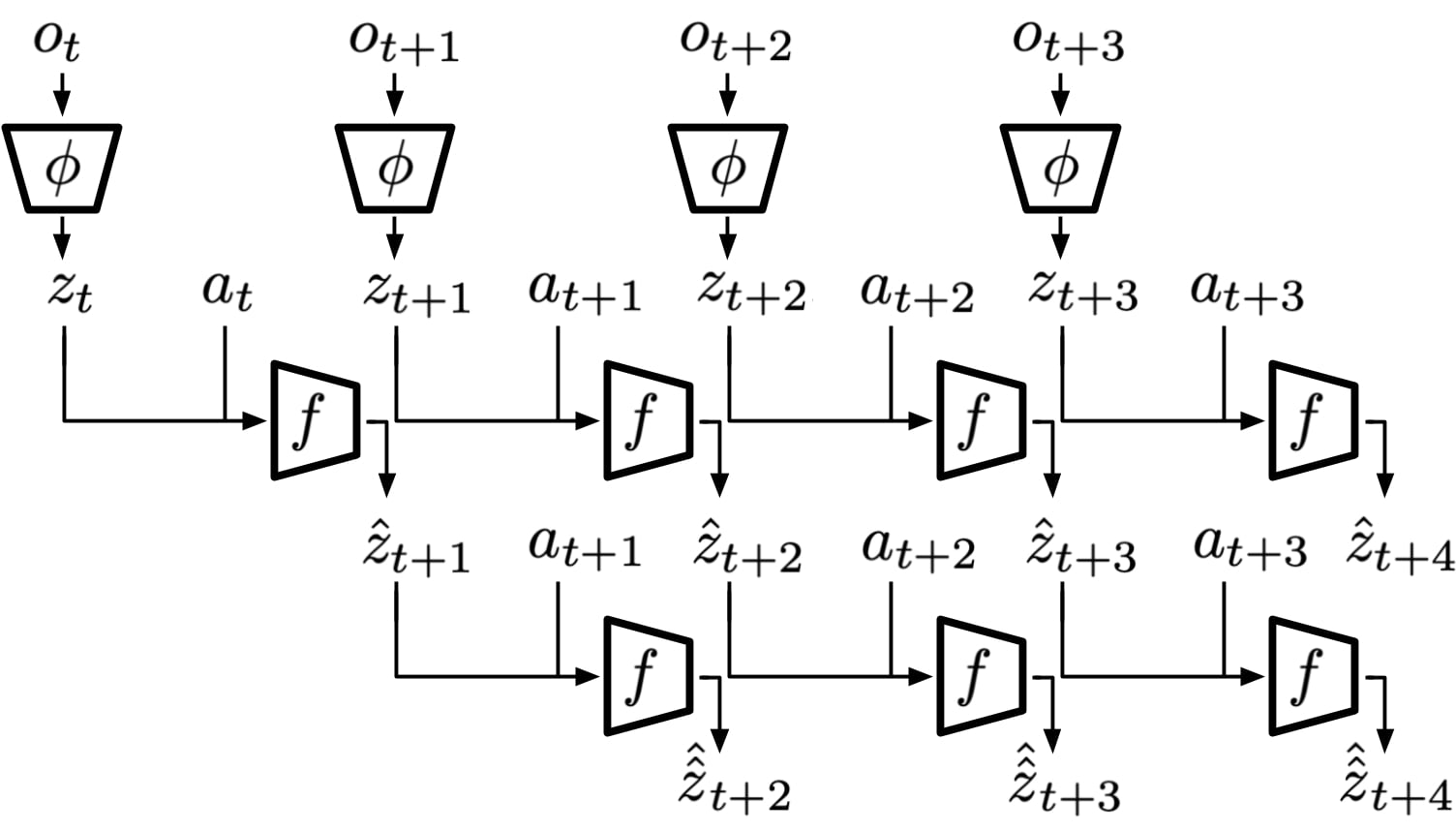}
  \label{fig:cfd_multi_step}
\end{subfigure}
\caption{Rollouts of the multi-step future predictions in the learned latent space. For instance, $\hat{z}_{t+2}$ and $\hat{\hat{z}}_{t+2}$ are one and two step predictions of $z_{t+2}$, respectively. In our experiments, we use 5 step prediction for a trajectory length of 32.}
\label{fig:cfd2}
\end{figure}
In practice, while forming the negative pairs we pick all the other latent observations in the same mini-batch, which also contains observations from the same sequence. To further enforce the prediction quality, we perform multi-step future predictions by continuously applying the forward dynamics model. These longer horizon predictions optimize the same objective given in Eq. \ref{eq:cfd} where $\hat{z}_i$ is replaced with any multi-step prediction of $z_i$. Fig. \ref{fig:cfd2} illustrates how multi-step predictions are obtained using a single forward dynamics model.

\section{Simulated and Real Environments and Tasks}
The primary manipulation task we have used in this work is vision-based stacking of one cube on top of another. However, as this is a particularly hard task to solve~\cite{riedmiller2018learning} from pixels from scratch with off-the-shelf RL algorithms, we studied the ablation effects of different components of our proposed RL framework on the easier problem of vision-based lifting instead. As lifting is an easier task, and a required skill towards achieving stacking, we focused on the latter for the rest of our experimental analysis in simulation and for all our real world evaluations.

Fig. \ref{fig:overview} shows our real robot setup, which is composed of a 7-DoF Sawyer robotic arm, a basket and two cubes. The agent receives the front left and right $64\times64$ RGB camera images as observations, shown in Fig.~\ref{fig:stacking}. The two cameras are positioned in a way that can help disambiguate 3D positions of the arm and the objects. In addition to these images, our observations also consist of the pose of the cameras, end-effector position and angle, and the gripper finger angle. The action space of the agent is 4D Cartesian velocity control of the end effector, with an additional action for actuating the gripper.
The real environment is modelled in simulation using the MuJoCo~\cite{mujoco} simulator. Fig \ref{fig:overview} also shows the simulated version of our environment. Unless mentioned otherwise, all of our policies are trained in simulation with domain randomization and a shaped reward functions.

The shaped reward function for lifting is a combination of reaching, touching and lifting rewards.
Let $d_{gripper}$ be the Euclidean distance of a target object from the pinch site of the end effector, and $h_t, h_o$ be the target height and object height from the ground in meters. Our reach reward is defined as ${r_{reach}=\mathds{1}(d_{gripper}<0.01)}$, where $\mathds{1}$ is the indicator function. In practice we use reward shaping with the Gaussian tolerance reward function as defined in the DeepMind Control Suite \cite{Tassa2018DeepMindCS}, with bounds $[0, 0.01]$ and a margin of $0.25$. Our touch reward $r_{touch}=\mathds{1}_{contact}$ is binary and provided by our simulator upon contact with the object. Our lift reward is
$r_{lift} = \mathds{1}(|h_t-h_o|<0.1\; \land\; h_o<0.01)$ and the final shaped version we use during training:
${r_{lift\_shaped} = r_{reach} + \frac{1+r_{touch}}{2} + r_{lift}}$.
As before, in practice the distance $|h_t-h_o|$ is passed through the same tolerance  function as above, with bounds $[0,0.1]$ and a margin of $15$. For stacking we now have a top and a bottom target objects with positions $\bf x_{t}, x_{b}$. If the cubes are in contact and on top of each other, the reward is $1$. Otherwise, we have additional shaping to aid with training. More specifically, if ${h_o({\bf x_t}) \le 0.025m}$ we revert to a normalized lift reward for the top object $r_{stack}=\frac{r_{lift\_shaped}({\bf x_t})}{5}$. Otherwise, ${r_{stack}=\frac{2+r_{reach}({\bf x_{b}})+\mathds{1}(h_o({\bf x_t})>h_o({\bf x_b}) \lor ||{\bf x_t}-{\bf x_b}||^2<0.07)}{5}}$, to account for bringing the cubes closer to each other. In practice we set $r_{reach}({\bf x_b})=1$ if it's greater than 0.75.
\begin{table}[t]
\begin{center}
\begin{tabular}{|l|l|c|}
\hline
\textbf{Training} & \textbf{Method} & \textbf{Task Success}\\
\hline
\hline
 &Domain Randomization & 46.0 \%  \\
End-to-End &DANN & 50.0 \%  \\
  &SSDA with TCN & 38.0 \%  \\
\hline
 &DANN & 50.0 \%  \\

Two-Stage  &SSDA with TCN  (Ours) & \textbf{54.0} \%  \\
 & SSDA with CFD (Ours) & \textbf{62.0} \%  \\
\hline
\end{tabular}
\end{center}
\caption{Sim-to-real transfer performance for vision-based cube stacking agent with unsupervised domain adaptation using DANN, self-supervised domain adaptation (SSDA) using TCN and CFD for the end-to-end and two-stage methods.}
\label{tbl:results}
\end{table}

\begin{table}[t]
\begin{center}
\begin{tabular}{|l|c|}
\hline
\textbf{Method} & \textbf{Task Success}\\
\hline
\hline
SSDA without Task Objective & 12.0 \%  \\
SSDA with Task Objective (Ours) & \textbf{62.0} \%  \\
\hline
\end{tabular}
\end{center}
\caption{Cube stacking performance on the real system for two-stage self-supervised domain adaptation (SSDA) with CFD optimized with and without the task objective.}
\label{tbl:results2}
\end{table}
In the real world, the cubes are fitted with AR tags that are only used for the purposes of fair and consistent evaluation of our resulting policies: \textbf{the 3D poses of the cubes are never available to an RL agent during training or testing}. At the beginning of every episode, the cubes are placed in a random position by a hand-crafted controller. All real world evaluations referred to in the rest of the section are on the stacking task and consist of 50 episodes. A real world episode is considered a success if the green cube is on top of the yellow cube at any point throughout the episode. Episodes are of length 200 with 20Hz control rate for both simulated and real environments.

\section{Experimental Results and Discussion}
\label{sec:exp_results}
In this section, we discuss the details of our experiments, and attempt to answer the following questions: \textsl{(a)} Can sequence-based self-supervision be used as a common auxiliary objective for simulated and real data without degrading task performance in simulation? \textsl{(b)} Does doing so improve final task performance in the real world? \textsl{(c)} How does using sequence-based self supervision for visual domain alignment between simulation and reality compare with domain-adversarial adaptation? \textsl{(d)} Is the use of actions in such a self-supervised loss important for bridging the sim-to-real domain gap? \textsl{(e)} What is the performance difference of modality tuning in our two-stage approach versus a one-stage end-to-end approach? and \textsl{(f)}  What are the effects of the different components of our RL framework in solving manipulation tasks from scratch, i.e. without the shared replay buffer or behavior cloning, in simulation?

\begin{figure}[t]
\centering
\begin{subfigure}{0.486\textwidth}
  \centering
  \includegraphics[width=0.95\linewidth]{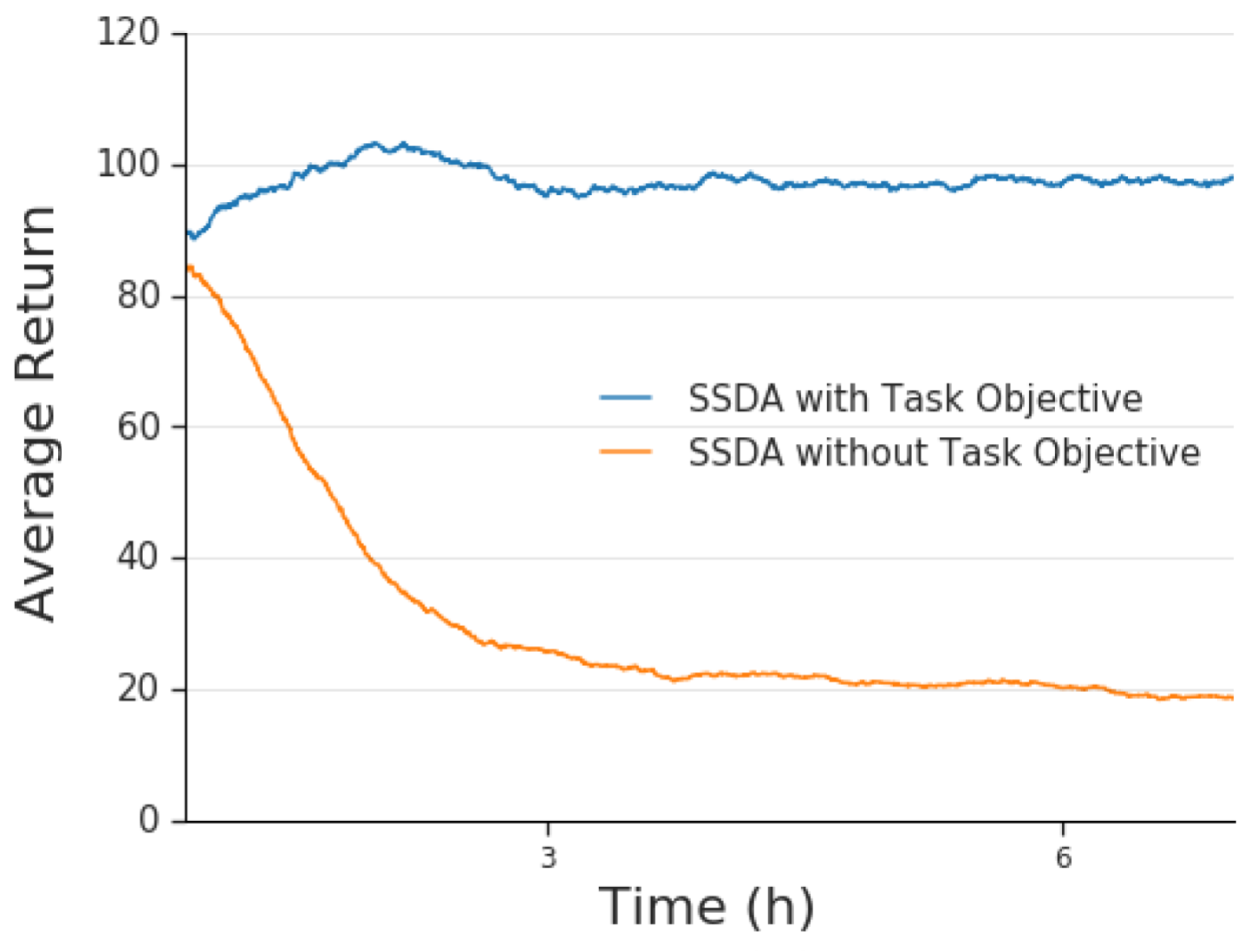}
  \label{fig:task_obj_abl}
\end{subfigure}
\caption{Cube stacking performance in simulation for two-stage self-supervised domain adaptation (SSDA) with CFD jointly optimized with and without the task objective.}
\label{fig:task_obj_abl}
\end{figure}

\subsection{Self-Supervised Sim-to-Real Adaptation}

We evaluated the following methods on our vision-based cube stacking task: domain randomization~\cite{sadeghi2016cad2rl}, unsupervised domain adaptation with a domain adversarial (DANN)~\cite{dann} loss, and self-supervised domain adaptation (SSDA) with two sequence-based self-supervised objectives: the time-contrastive networks (TCN)~\cite{tcn} loss, and the contrastive forward dynamics (CFD)
loss we proposed in Sect.~\ref{sec:cd}. We ablate two different training methods for domain adaptation, end-to-end and two-stage. The end-to-end training method simply optimize Eq. \ref{eq:ss} from Sect. \ref{sec:selfsup} with respect to all parameters, without the two-stage procedure described in Sect. \ref{sec:selfsup}.
This means that all of the losses are jointly optimized without freezing any part of the neural network. Two-stage training procedure is described in Sect. \ref{sec:our_methods} and employs modality tuning~\cite{aytar2017cross}.

Table \ref{tbl:results} shows the quantitative results from evaluating task success on the real robot. These experiments show that DANN improves on top of the domain randomization baseline by a small margin. However, end-to-end adaptation with the TCN loss results in degradation of performance. This is likely due to insufficient sharing of the encoder between the self-supervised objective using simulated data and real data. On the other hand, the two-stage self-supervised domain adaptation with TCN significantly improves over the end-to-end variant and domain randomization baselines. This reconfirms that modality tuning used in the two-stage training method results in significantly better sharing of the encoder. Finally, the two-stage self-supervised adaptation with our CFD objective, which utilizes both the temporal structure of the observations and the actions, performs significantly better when compared to all other methods, yielding a 62 \% task success.

We also evaluated the importance of jointly optimizing the RL and BC objectives in Eq. \ref{eq:ss} for the two-stage self-supervised domain adaptation. As one can see in Table \ref{tbl:results2}, only optimizing $\mathcal{L}_{SS}$ without the task objective significantly reduces the performance. Fig.~\ref{fig:task_obj_abl} further shows how the task performance in simulation degrades when optimizing only the self-supervised objective. In essence, by only optimizing the self-supervised loss, the network \emph{catastrophically forgets}~\cite{forget} how to solve the manipulation task.
\begin{figure}[t]
\centering
\begin{subfigure}{0.489\textwidth}
  \centering
  \includegraphics[width=0.95\linewidth]{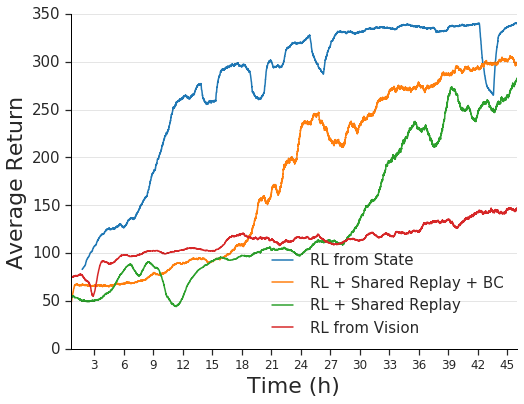}
  \label{fig:shared_replay_abl}
\end{subfigure}
\caption{Ablation of techniques used in conjunction with RL for cube lifting task in simulation. The plot shows the average return for the lifting task with and without shared replay buffer and behavior cloning (BC). RL from state and RL from vision are trained only with the RL objective.}
\label{fig:shared_replay_abl}
\end{figure}
\subsection{Ablations for different components of our RL framework}
In order to assess the necessity and efficacy of the different components of our framework, described in Sect.~\ref{sec:aai}, we provide ablation experimental results. Specifically we examined the effects of the state-based agent that share a replay buffer with the vision-based agent, and the addition of an auxiliary behavior cloning objective for the vision-based agent to imitate the state-based agent.  Fig.~\ref{fig:shared_replay_abl} shows these effects on the cube lifting task. A vision-based agent trained with MPO~\cite{mpo}, the state-of-the-art continuous control RL method at the core of our framework, struggles with solving this task, contrary to an MPO agent with access to the full state information. By sharing the replay buffer between the state-based agent and the vision-based agent, one can see that the vision-based agent is able to solve lifting in a reasonable amount of time. The addition of the behavior cloning (BC) objective further improves the speed and stability of training.

Fig.~\ref{fig:bc_abl} shows the even more profound effect our BC objective has on learning our vision-based cube stacking task. Furthermore, one can also observe the stability of the method persists even when jointly training, end-to-end, with the TCN loss, or the DANN loss with real world data.

\begin{figure}[t]
\centering
\begin{subfigure}{0.486\textwidth}
  \centering
  \includegraphics[width=0.95\linewidth]{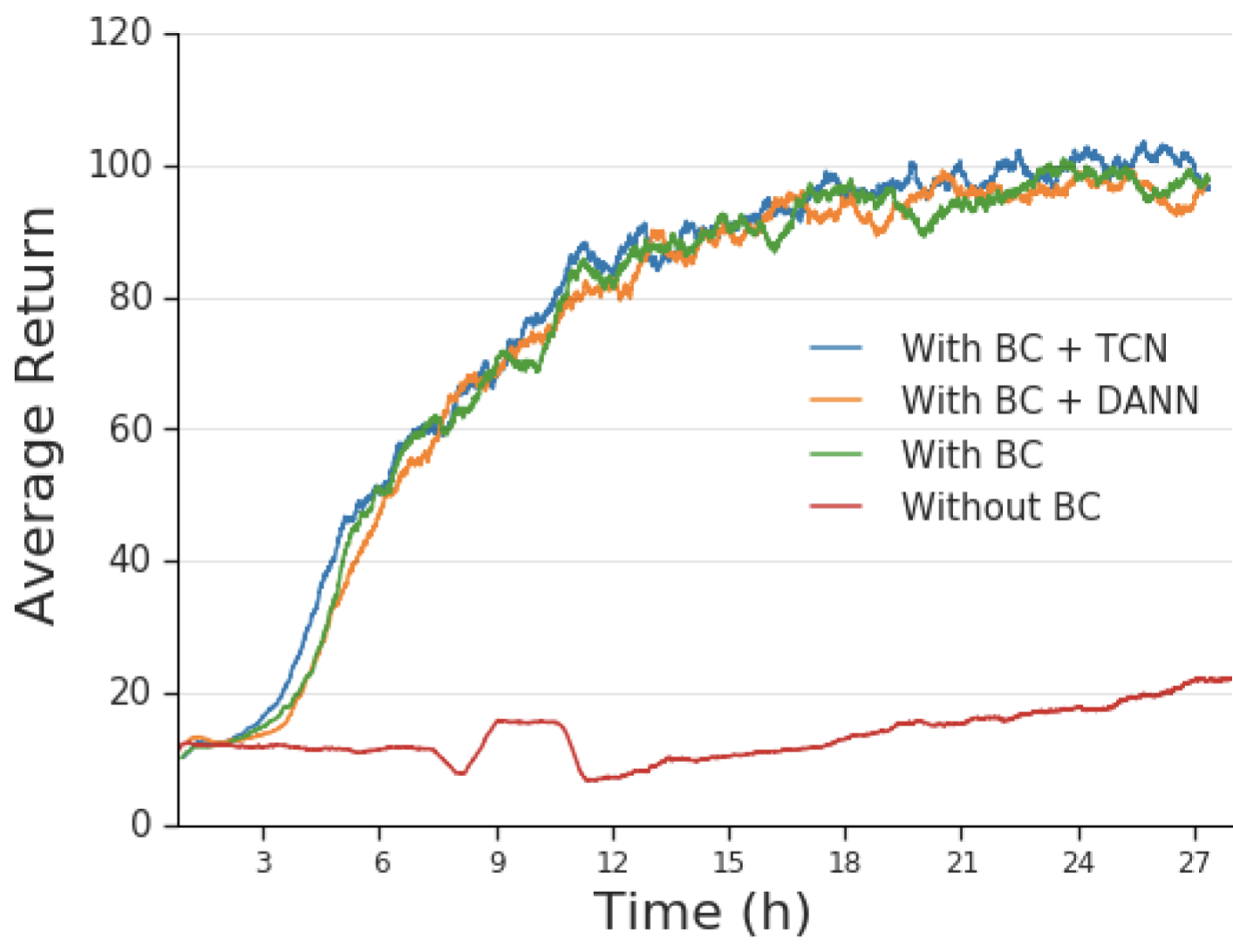}
  \label{fig:bc_abl}
\end{subfigure}
\caption{Simulation performance on our vision-based stacking task of our RL framework with and without behavior cloning (BC). Using BC results in faster training that maintains stability with the addition of auxiliary adaptation objectives.}
\label{fig:bc_abl}
\end{figure}

\section{Conclusion}
\label{sec:conclusion}
In this work, we have presented our self-supervised domain adaptation method, which uses unlabeled real robot data to improve sim-to-real transfer learning. Our method is able to perform domain adaptation for sim-to-real transfer learning of cube stacking from visual observations. In addition to our domain adaptation method, we developed \textit{contrastive forward dynamics} (CFD), which combines dynamics model learning with time-contrastive techniques to better utilize the structure available in unlabeled robot data. We demonstrate that using our CFD objective for adaptation yields a clear improvement over domain randomization, other self-supervised adaptation techniques and domain adversarial methods.

Through our experiments, we discovered that optimizing only the first visual layers of the policy network in combination with jointly optimizing the reinforcement learning, behavior cloning and self-supervised loss was necessary for a successful application of self-supervised learning for sim-to-real transfer for robotic manipulation. Finally, the use of sequence-based self-supervised loss by leveraging the dynamical structure in the robotic system ultimately resulted in the best domain adaptation for our manipulation task.

\addtolength{\textheight}{0cm}   
\bibliographystyle{IEEEtran}
\bibliography{IEEEabrv,reference}

\end{document}